\documentclass[11pt,a4paper]{article}
\usepackage[hyperref]{emnlp-ijcnlp-2019}
\usepackage{times}
\usepackage{latexsym}

\usepackage{url}

\usepackage{cite}
\usepackage{amsmath,amssymb,amsfonts}
\usepackage{algorithmic}
\usepackage{graphicx}
\usepackage{textcomp}
\usepackage{xcolor}
\usepackage{booktabs}
\usepackage{stfloats}

\aclfinalcopy

\definecolor{st}{rgb}{0.1118, 0.60, 0.2235}

\definecolor{gs}{rgb}{0.1118, 0.2235, 0.60}

\newcommand{\myeq}{\mkern5mu{=}\mkern5mu}

\title{Behavior Gated Language Models}

\author{Prashanth Gurunath Shivakumar\thanks{~~These authors contributed equally to this work} \and Shao-Yen Tseng\footnotemark[1] \\ {\bf Panayiotis Georgiou} \and {\bf Shrikanth Narayanan}\\
  Signal Analysis and Interpretation Laboratory \\
  Department of Electrical and Computer Engineering \\
  University of Southern California \\
  Los Angeles, CA, USA \\
  {\tt pgurunat@usc.edu, shaoyent@usc.edu},\\{\tt georgiou@sipi.usc.edu, shri@sipi.usc.edu}
}

\date{}
\begin{document}

\maketitle

\begin{abstract}
Most current language modeling techniques only exploit co-occurrence, semantic and syntactic information from the sequence of words.
However, a range of information such as the state of the speaker and dynamics of the interaction might be useful.
In this work we derive motivation from psycholinguistics and propose the addition of behavioral information into the context of language modeling.
We propose the augmentation of language models with an additional module which analyzes the behavioral state of the current context.
This behavioral information is used to gate the outputs of the language model before the final word prediction output.
We show that the addition of behavioral context in language models achieves lower perplexities on behavior-rich datasets.
We also confirm the validity of the proposed models on a variety of model architectures and improve on previous state-of-the-art models with generic domain Penn Treebank Corpus.

\end{abstract}

\section{Introduction}

Recurrent neural network language models (RNNLM) can theoretically model the word history over an arbitrarily
long length of time and thus have been shown to perform better than traditional n-gram models
\citep{mikolov2010recurrent}.
Recent prior work has continuously improved the performance of RNNLMs through hyper-parameter
tuning, training optimization methods, and development of new network architectures
\citep{zaremba2014recurrent, merity2017regularizing, bai2018trellis, dai2019transformer}.

On the other hand, many work have proposed the use of domain knowledge and additional information
such as topics or parts-of-speech to improve language models. 
While syntactic tendencies can be inferred from a few preceding words, semantic coherence may
require longer context and high level understanding of natural language, both of which are difficult to
learn through purely statistical methods. 
This problem can be overcome by exploiting external information to capture long-range semantic
dependencies. 
One common way of achieving this is by incorporating part-of-speech (POS) tags into the RNNLM as
an additional feature to predict the next word \citep{gong2014recurrent, su2017parallel}.
Other useful linguistic features include conversation-type, which was shown to improve language modeling when combined with POS tags \citep{shi2010language}.
Further improvements were achieved through the addition of socio-situational setting information and other linguistic features such as lemmas and topic \citep{shi2012towards}. 

The use of topic information to provide semantic context to language models has also been
studied extensively \citep{mikolov2012context, ghosh2016contextual, dieng2017topicrnn,
wang2018topic}.
Topic models are useful for extracting high level semantic structure via latent topics which can aid
in better modeling of longer documents.


Recently, however, empirical studies involving investigation of different network architectures,
hyper-parameter tuning, and optimization techniques have yielded better performance than the addition
of contextual information \citep{press2019partially, krause2019dynamic}.
In contrast to the majority of work that focus on improving the neural network aspects of RNNLM, 
we introduce psycholinguistic signals along with linguistic units to improve the fundamental language model. 

In this work, we utilize behavioral information embedded in the language to aid the language model.
We hypothesize that different psychological behavior states incite differences in the use of
language \citep{pennebaker2001patterns,lindquist2015role}, and thus modeling these tendencies can provide useful information in statistical language
modeling.
And although not directly related, behavioral information may also correlate with conversation-type
and topic. 
Thus, we propose the use of psycholinguistic behavior signals as a gating mechanism to augment typical
language models.
Effectively inferring behaviors from sources like spoken text, written articles can lead to
personification of the language models in the speaker-writer arena.


\section{Methodology}
In this section, we first describe a typical RNN based language model which serves as a baseline for
this study.
Second, we introduce the proposed behavior prediction model for extracting behavioral information.
Finally, the proposed architecture of the language model which incorporates the
behavioral information through a gating mechanism is presented.

\subsection{Language Model}
\begin{figure}[t]
\centering
\includegraphics[width=0.8\linewidth]{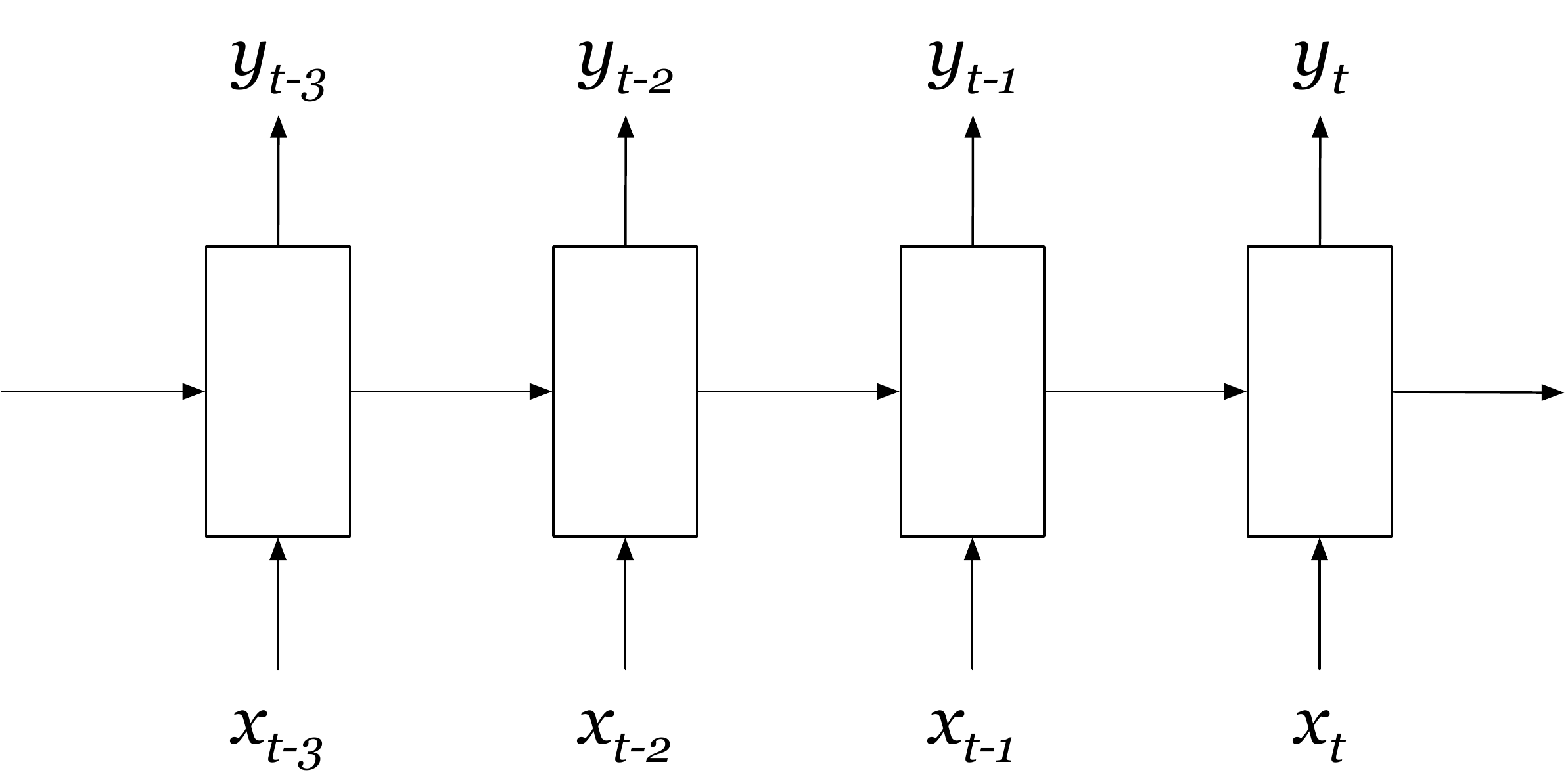}
\caption{RNN language model.} \label{fig:rnn_lm}
\vspace{-1em}
\end{figure}

The basic RNNLM consists of a vanilla unidirectional LSTM which predicts the next word given the
current and its word history at each time step.
In other words, given a sequence of words $ \mathbf{x} \myeq x_1, x_2, \ldots x_n$ as input, the network predicts
a probability distribution of the next word $ y $ as 
$ P(y \mid \mathbf{x}) $. 
Figure~\ref{fig:rnn_lm} illustrates the basic architecture of the RNNLM.

Since our contribution is towards introducing behavior as a psycholinguistic feature for aiding the
language modeling process, we stick with a reliable and simple LSTM-based RNN model and
follow the recommendations from \citet{zaremba2014recurrent} for our baseline model.

\subsection{Behavior Model} \label{sec:beh_model}
The analysis and processing of human behavior informatics is crucial in many psychotherapy settings
such as observational studies and patient therapy \citep{narayanan2013behavioral}.
Prior work has proposed the application of neural networks in modeling human behavior in a variety
of clinical settings \citep{Xiao2016BehavioralCodingofTherapist, tseng2016couples,
gibson2017attentionnetworks}.

In this work we adopt a behavior model that predicts the likelihood of occurrence of various
behaviors based on input text. 
Our model is based on the RNN architecture in Figure~\ref{fig:rnn_lm}, but instead of the next word
we predict the joint probability of behavior occurrences $ P(\mathbf{B} \mid \mathbf{x}) $ where
$ \mathbf{B} \myeq \{b_{i}\}$ and $ b_{i} $
is the occurrence of behavior $i$.
In this work we apply the behaviors: \textit{Acceptance}, \textit{Blame}, \textit{Negativity},
\textit{Positivity}, and \textit{Sadness}.
This is elaborated more on in Section~\ref{sec:setup}. 

\subsection{Behavior Gated Language Model}
\subsubsection{Motivation}
Behavior understanding encapsulates a long-term trajectory of a person's psychological state. 
Through the course of communication, these states may manifest as short-term instances of emotion or
sentiment.
Previous work has studied the links between these psychological states and their effect on 
vocabulary and choice of words \citep{pennebaker2001patterns} as well as use of language \citep{lindquist2015role}.
Motivated from these studies, we hypothesize that due to the duality of behavior and language we can
improve language models by capturing variability in language use caused by different
psychological states through the inclusion of behavioral information.

\begin{figure}[t]
\centering
\includegraphics[width=0.7\linewidth]{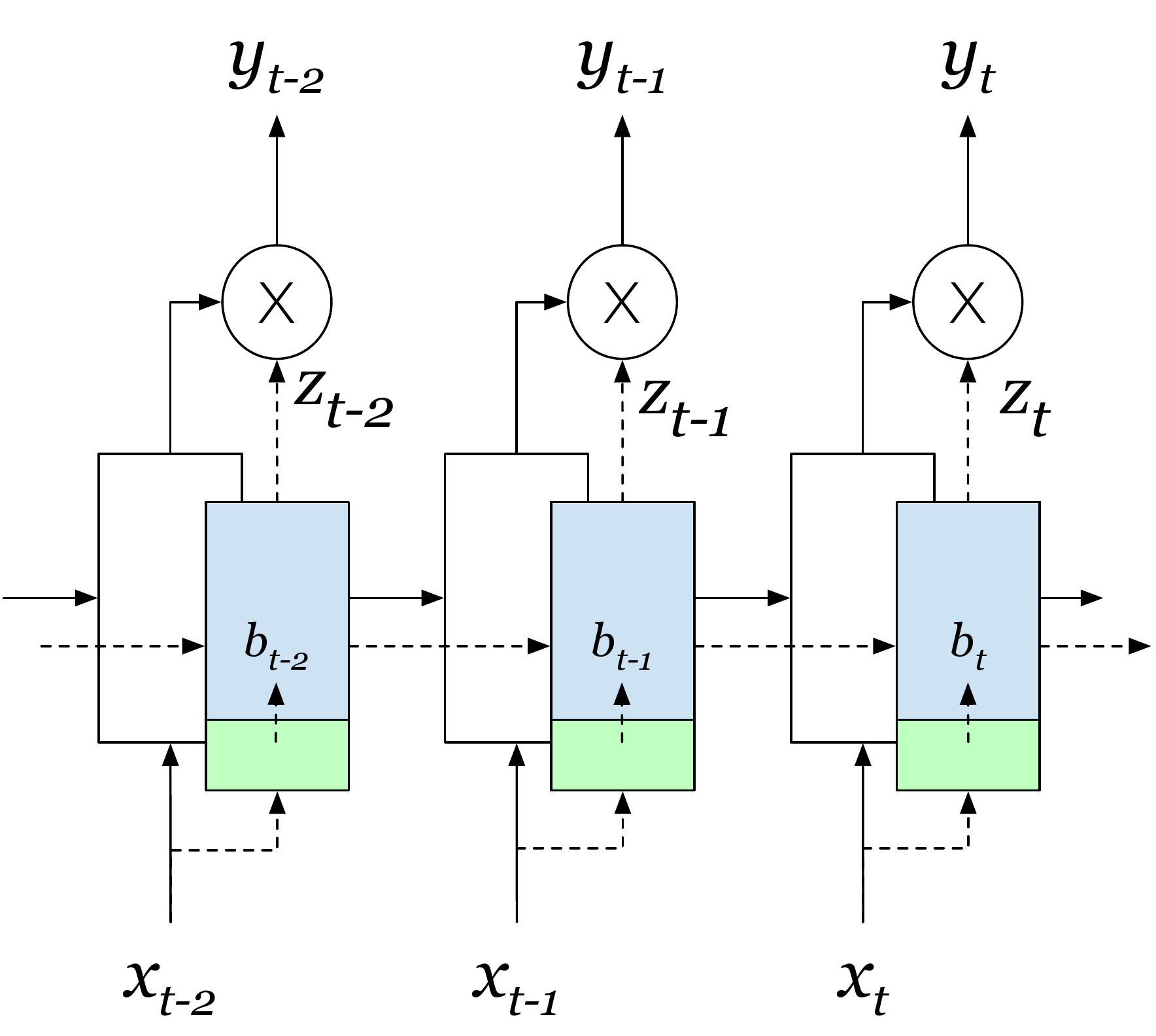}
\caption{Behavior gated language model.}\label{fig:beh_gated_lm}
\vspace{-1em}
\end{figure}

\subsubsection{Proposed Model}

We propose to augment RNN language models with a behavior model that provides information relating
to a speaker's psychological state. 
This behavioral information is combined with hidden layers of the RNNLM through a gating mechanism
prior to output prediction of the next word. 
In contrast to typical language models, we propose to model $ P(\mathbf{y} \mid \mathbf{x}, \mathbf{z}) $ where $ \mathbf{z} \equiv f( P(\mathbf{B}\mid \mathbf{x}))$ for an RNN function $f(\cdot)$. 
The behavior model is implemented with a multi-layered RNN over the input sequence of words. 
The first recurrent layer of the behavior model is initialized with pre-trained weights
from the model described in Section~\ref{sec:beh_model} and fixed during language modeling training. 
An overview of the proposed behavior gated language model is shown in Figure~\ref{fig:beh_gated_lm}. 
The RNN units shaded in green (lower section) denote the pre-trained weights from the behavior classification model which are fixed during the entirety of training.
The abstract behavior outputs $ b_t $ of the pre-trained model are fed into a time-synced RNN, denoted in blue (upper section), which is subsequently used for gating the RNNLM predictions.
The un-shaded RNN units correspond to typical RNNLM and operate in parallel to the former.


\section{Experimental Setup} \label{sec:setup}
\subsection{Data} \label{sec:data}
\subsubsection{Behavior Related Corpora}
For evaluating the proposed model on behavior related data, we employ the Couples Therapy Corpus
(CoupTher) \citep{christensen2004traditional} and Cancer Couples Interaction Dataset (Cancer)
\citep{reblin2018cancer}.
These are the targeted conditions under which a behavior-gated language model can offer improved performance. 

\noindent\textbf{Couples Therapy Corpus:} This corpus comprises of dyadic conversations between real couples seeking marital counseling.
The dataset consists of audio, video recordings along with their transcriptions.
Each speaker is rated by multiple annotators over 33 behaviors.
The dataset comprises of approximately 0.83 million words with 10,000 unique entries of which 0.5 million is used for training (0.24m for dev and 88k for test).

\noindent\textbf{Cancer Couples Interaction Dataset:}
This dataset was gathered as part of a observational
study of couples coping with advanced cancer.
Advanced cancer patients and their spouse caregivers
were recruited from clinics and asked to interact with each other in two
structured discussions: neutral discussion and cancer related.
Interactions were audio-recorded using
small digital recorders worn by each participant.
Manually transcribed audio has approximately 230,000 word tokens with a vocabulary size of 8173.

\subsubsection{Penn Tree Bank Corpus}
In order to evaluate our proposed model on more generic language modeling tasks, we employ Penn Tree bank (PTB) \citep{marcus1994penn}, as preprocessed by \citet{mikolov2011empirical}.
Since Penn Tree bank mainly comprises of articles from Wall Street Journal it is not expected to contain substantial expressions of behavior.


\subsection{Behavior Model}

The behavior model was implemented using an RNN with LSTM units and trained with the Couples Therapy Corpus.
Out of the 33 behavioral codes included in the corpus we applied the behaviors \textit{Acceptance}, \textit{Blame}, \textit{Negativity}, \textit{Positivity}, and \textit{Sadness} to train our models.
This is motivated from previous works which showed good separability in these behaviors as well as
being easy to interpret.
The behavior model is pre-trained to identify the presence of each behavior from a sequence of words
using a multi-label classification scheme. 
The pre-trained portion of the behavior model was implemented using a single layer RNN with LSTM
units with dimension size 50. 

\subsection{Hyperparameters}

We augmented previous RNN language model architectures by \citet{zaremba2014recurrent} and \citet{merity2017regularizing} with our proposed behavior gates.
We used the same architecture as in each work to maintain similar number of parameters and performed a grid search of hyperparameters such as learning rate, dropout, and batch size. 
The number of layers and size of the final layers of the behavior model was also optimized. 
We report the results of models based on the best validation result.

\section{Results}
We split the results into two parts. We first validate the proposed technique on behavior related language modeling tasks and then apply it on more generic domain Penn Tree bank dataset.

\subsection{Behavior Related Corpora}

\subsubsection{Couple's Therapy Corpus}
We utilize the Couple's Therapy Corpus as an in-domain experimental corpus since our behavior classification model is also trained on the same.
The RNNLM architecture is similar to \citet{zaremba2014recurrent}, but with hyperparameters optimized for the couple's corpus.
The results are tabulated in Table~\ref{tab:res_couples} in terms of perplexity.
We find that the behavior gated language models yield lower perplexity compared to vanilla LSTM language model.
A relative improvement of 2.43\% is obtained with behavior gating on the couple's data.

\subsubsection{Cancer Couples Interaction Dataset}
To evaluate the validity of the proposed method on an out-of-domain but behavior related task, we utilize the Cancer Couples Interaction Dataset.
Here both the language and the behavior models are trained on the Couple's Therapy Corpus.
The Cancer dataset is used only for development (hyper-parameter tuning) and testing.
We observe that the behavior gating helps achieve lower perplexity values with a relative improvement of 6.81\%.
The performance improvements on an out-of-domain task emphasizes the effectiveness of behavior gated language models.

\begin{table}[t]
        \centering
        \begin{tabular}{l | c c}
                \toprule
                \textbf{Model} & \textbf{CoupTher} & \textbf{Cancer} \\
                \midrule
                LSTM & 66.32 &	159.65 \\
                ~~~ + Behavior gating & 64.71 & 148.78  \\
                
                \bottomrule
        \end{tabular}
        \caption{Language model test perplexities on Couples Therapy and Cancer Couples Interaction Dataset.}
        \label{tab:res_couples}
        \vspace{-1em}
\end{table}

\begin{table*}[!b]
        \centering
        \begin{tabular}{l | c c c}
                \toprule
                \textbf{Model} & \textbf{\# Params} & \textbf{Validation} & \textbf{Test} \\
                \midrule
                LSTM-Medium \citep {zaremba2014recurrent} & 20M & 86.2 & 82.7 \\
                ~~~ + Behavior gating & 20M & 83.85 & 78.75 \\
                \midrule
                LSTM-Large \citep{zaremba2014recurrent} & 66M & 82.2 & 78.4  \\
                ~~~ + Behavior gating & 67M & 80.09 & 75.80 \\
                \midrule
                AWD-LSTM \citep{merity2017regularizing} & 24M & 60.0 & 57.3 \\
                ~~~ + Behavior gating & 27M & 59.15 & 56.92 \\
                \bottomrule
        \end{tabular}
        \caption{Language model perplexities on validation and test sets of Penn Treebank.}
        \label{tab:res_ptb}
\end{table*}

\subsection{Penn Tree Bank Corpus}
Although the proposed model is motivated and targeted towards behavior related datasets, the hypothesis should theoretically extend towards any human generated corpora.
To assess this, we also train models on a non-behavior-rich database, the Penn Tree Bank Corpus.
We experiment with both the \textit{medium} and \textit{large} architectures proposed by \citet{zaremba2014recurrent}.
The perplexity results on PTB are presented in Table \ref{tab:res_ptb}.
All language models showed an improvement in perplexity through the addition of behavior gates. 
It can also be observed that LSTM-Medium with behavior gating gives similar performance to baseline LSTM-Large even though the latter has more than three times the number of parameters. 

\subsubsection{Previous state-of-the-art architectures}
Finally we apply behavior gating on a previous state-of-the-art architecture, one that is most often used as a benchmark over various recent works.
Specifically, we employ the AWD-LSTM proposed by \citet{merity2017regularizing} with QRNN \citep{qrnn} instead of LSTM.
We observe positive results with AWD-LSTM augmented with behavior-gating providing a relative improvement of (1.42\% on valid) 0.66\% in perplexity (Table~\ref{tab:res_ptb}).

\section{Conclusion \& Future Work}
In this study, we introduce the state of the speaker/author into language modeling in the form of behavior signals.
We track 5 behaviors namely \textit{acceptance}, \textit{blame}, \textit{negativity}, \textit{positivity} and \textit{sadness} using a 5 class multi-label behavior classification model.
The behavior states are used as gating mechanism for a typical RNN based language model.
We show through our experiments that the proposed technique improves language modeling perplexity specifically in the case of behavior-rich scenarios.
Finally, we show improvements on the previous state-of-the-art benchmark model with Penn Tree Bank Corpus to underline the affect of behavior states in language modeling.

In future, we plan to incorporate the behavior-gated language model into the task of automatic speech recognition (ASR).
In such scenario, we could derive both the past and the future behavior states from the ASR which could then be used to gate the language model using two pass re-scoring strategies.
We expect the behavior states to be less prone to errors made by ASR over a sufficiently long context and hence believe the future behavior states to provide further improvements.



\bibliographystyle{acl_natbib}
\bibliography{ref} 

\end{document}